\documentclass{vgtc}                          

\graphicspath{{figures/}{pictures/}{images/}{./}}

\usepackage{times}

\usepackage{mathptmx}

\usepackage{graphicx}
\usepackage{adjustbox}
\usepackage{float}
\usepackage{stfloats}
\usepackage[percent]{overpic}
\usepackage{subcaption}
\usepackage{caption}

\usepackage{booktabs}
\usepackage{multirow}
\usepackage{longtable}
\usepackage{array}
\usepackage{multicol}

\usepackage{amsmath}
\usepackage{amssymb}
\usepackage{amsfonts}
\usepackage{optidef}

\usepackage[ruled,vlined,linesnumbered,boxed]{algorithm2e}
\usepackage{algpseudocode}

\usepackage{textcomp}
\usepackage{xcolor}
\usepackage{verbatim}

\usepackage{breakurl}
\usepackage{xurl}

\onlineid{0}
\vgtccategory{Research}
\vgtcinsertpkg

\title{Understanding Sensor Vulnerabilities in Industrial XR Tracking}

\author{
Sourya Saha\textsuperscript{*}\thanks{Both authors contributed equally to this work.}\thanks{e-mail: ssaha2@gradcenter.cuny.edu}\\ %
\scriptsize City University of New York %
\and
Md.\ Nurul Absur\textsuperscript{*}\thanks{e-mail: mabsur@gradcenter.cuny.edu}\\ %
\scriptsize City University of New York %
}

\abstract{
Extended Reality (XR) systems deployed in industrial and operational settings rely on Visual--Inertial Odometry (VIO) for continuous six-degree-of-freedom pose tracking, yet these environments often involve sensing conditions that deviate from ideal assumptions. Despite this, most VIO evaluations emphasize nominal sensor behavior, leaving the effects of sustained sensor degradation under operational conditions insufficiently understood. This paper presents a controlled empirical study of VIO behavior under degraded sensing, examining faults affecting visual and inertial modalities across a range of operating regimes. Through systematic fault injection and quantitative evaluation, we observe a pronounced asymmetry in fault impact where degradations affecting visual sensing typically lead to bounded pose errors on the order of centimeters, whereas degradations affecting inertial sensing can induce substantially larger trajectory deviations, in some cases reaching hundreds to thousands of meters. These observations motivate greater emphasis on inertial reliability in the evaluation and design of XR systems for real-life industrial settings.
}


\begin{document}

\firstsection{Introduction}
\maketitle
\label{sec:intro}

Extended Reality (XR) is showing great promise in industrial environments, including manufacturing, equipment maintenance, safety measurement, remote assistance, and worker training. These industrial applications heavily rely on precise, continuous six-degree-of-freedom (6-DoF) pose tracking to ensure alignment between virtual content and the physical world \cite{ESWARAN2026130877, 10941631}. In most advanced XR systems, inside-out tracking based on Visual-Intertial Odometry (VIO) has become the prevailing approach, combining camera parameters with inertial sensors to estimate device motion in near real time \cite{10.1145/3652595}. Although VIO performs well in controlled environments, real-life industrial settings face significantly harsher operating conditions, including anomalies in machine systems, temporary occlusions, dust accumulation, lighting changes, and electromagnetic interference \cite{10.1145/3704413.3765307}. These hindrance factors significantly affect sensor reliability and can compromise the robustness of pose-tracking systems used in real-world industrial XR workflows.

\begin{figure}[!t]
\centering
\includegraphics[width=\columnwidth]{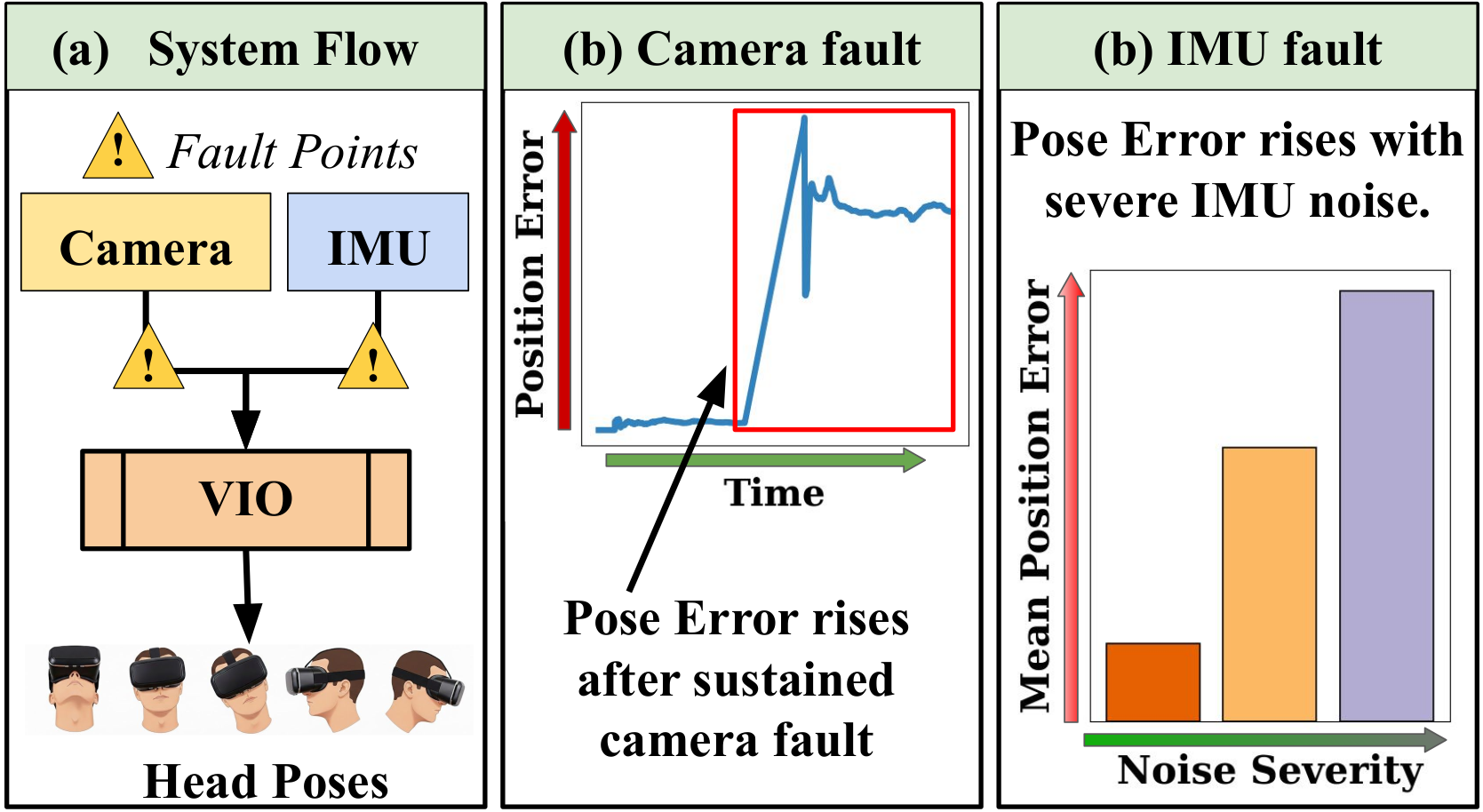}
\caption{Illustration of sensor fault behavior in XR VIO. (a) Conceptual pipeline showing fault points. (b) Pose error growth over time under sustained camera degradation. (c) Mean pose error as a function of IMU noise severity, highlighting pronounced error growth under severe inertial corruption.}
\label{fig:fig1}
\end{figure}

In real industrial XR deployments, sensor faults are not rare but an expected part of operations. Cameras may experience temporary occlusions due to dust, debris, or worker motion, while abrupt lighting changes can degrade visual feature tracking. Inertial sensors are similarly susceptible to vibration, electromagnetic interference, aging, and transient dropout. Despite the prevalence of these conditions, most evaluations of Visual-Inertial Odometry systems emphasize performance under nominal sensor operation, leaving the impact of sustained camera and IMU faults insufficiently examined. This makes it difficult to design effective sensor redundancy, fault detection, and degradation strategies for industrial XR systems \cite{SHARMA20232029}. Moreover, because VIO relies on tightly coupled fusion of visual and inertial measurements that play distinct roles in motion estimation, faults in different sensors can lead to markedly different error behaviors, with recovery influenced by fault type, duration, severity, and timing along the trajectory.

Existing VIO literature for XR is mostly developed and evaluated under the assumption of marginal sensing conditions, with emphasis on tracking accuracy, robustness, and low latency. Recent works on XR-oriented VIO systems focus on improving visual-inertial fusion, initialization, and feature tracking to better support mobile AR and VR scenarios \cite{xr-vio}. Recent findings in commercial XR devices highlight the high sensitivity of environmental and kinematic factors, such as lighting variation, feature density, motion blur, and user motion dynamics, in challenging industrial conditions. This degrades inside-out tracking performance and has an erroneous impact on the overall pipeline. In parallel, robustness-focused VIO methods aim to improve performance in visually degraded or dynamic environments by mitigating outliers, handling degenerate motion, and imposing inertial constraints \cite{enhancing-vio, rd-vio}. Although these advanced contributions show great promise, existing evaluations fail to account for sensor unavailability over time or capture degradation characteristics. Also, the current state of the art focuses on curated, \emph{clean} datasets such as \textit{EuRoC} \cite{euroc}
and \textit{TUM-VI} \cite{klenk2021tumvietumstereovisualinertial}, but they lack realistic scenario representation. \cref{fig:fig1} conceptually illustrates this gap by showing how sustained faults introduced at different points in the XR VIO sensing pipeline result in markedly different pose-error evolution and long-term tracking behavior. This motivates a systematic study of sensor fault resilience in XR tracking systems.

In this paper, we present a systematic study of sensor fault resilience in Visual-Inertial Odometry systems used for XR tracking. Specifically, we make the following contributions.
\begin{itemize}
\item We provide a structured characterization of how sustained sensor degradation affects XR VIO behavior beyond nominal operating conditions, explicitly considering faults that persist longer than transient measurement corruption.
\item We analyze and compare the impact of camera and inertial sensor faults across a broad range of fault characteristics, including variations in modality, severity, and temporal placement, to expose differences in fault sensitivity at the system level.
\item We conduct an extensive empirical evaluation under realistic XR motion to quantify how sensor faults influence pose accuracy and long-term tracking stability across sensing modalities.
\end{itemize}
Together, these contributions establish sensor fault resilience as a first-class concern for the design and evaluation of VIO-based XR tracking systems.

Our extensive evaluation is conducted on an XR-oriented VIO setup built on the ILLIXR runtime \cite{illixr}, a modular end-to-end platform for experimenting with XR system components under realistic execution conditions. The evaluation compares fault-affected operation against nominal sensing to isolate the impact of camera and inertial degradation on tracking quality. Pose error is used as a high-level measure to capture both short-term deviations during fault exposure and longer-term drift over extended motion. Across the evaluated scenarios, inertial sensor degradation produces substantially larger pose errors than visual degradation, with representative position errors ranging from hundreds to nearly 4,000 m. These results provide a system-level view of fault resilience in XR VIO and highlight the importance of accounting for sensor reliability when evaluating XR tracking performance.

The remainder of this paper is organized as follows. Section \cref{sec:related-works} reviews prior work on VIO and fault-aware estimation approaches relevant to XR tracking systems. Section \cref{sec:framework} describes the empirical framework used to study sensor fault resilience in XR VIO. Section \cref{sec:expertimental} presents the experimental analysis and evaluation results. 
Finally, Section \cref{sec:conclusion} concludes the paper, discusses some implications of the results, states some limitations, and outlines directions for future work.

\section{Related Works}
\label{sec:related-works}

VIO has been widely studied in robotics and AR/VR, leading to diverse works on geometry-based and learning-augmented methods that improve pose estimation accuracy under nominal or mildly degraded sensing conditions. The advanced MSCKF-based VIO algorithm \cite{du2024pomsckfefficientvisualinertialodometry} shows great results and does not require any feature position information. In \cite{orbslam3, eqvio}, geometry-driven approaches, including optimization-based visual and visual-inertial SLAM systems, are evaluated in various VIO pipelines. In recent work, such as \cite{viodualpronet}, neural components have been introduced into classical estimation frameworks to improve robustness to transient disturbances, including time-varying sensor noise and degraded visual inputs. In parallel, hybrid approaches combine traditional geometric visual tracking with learned feature representations to enhance performance under challenging visual environments \cite{mixvio}. Although the aforementioned studies show great promise, existing VIO methods focus on accuracy under nominal conditions and are tested in clean benchmarks, which lack systematic characterization under prolonged or severe sensor faults.

Recent advances in VIO have led to several mature estimation pipelines that serve as tracking backbones in real-time perception and tracking systems. Prior work has explored modular visual–inertial architectures that emphasize real-time operation and integration within larger perception stacks \cite{rosinol2020kimeraopensourcelibraryrealtime}, as well as scalable keyframe-based optimization frameworks that balance estimation accuracy with computational efficiency \cite{okvis2x}. Other approaches improve robustness in visually challenging environments by incorporating richer geometric cues, such as combining point and line features within tightly coupled visual–inertial formulations \cite{He2018PLVIOTM}, while related extensions introduce additional sensing constraints to reduce long-term drift over extended trajectories \cite{cao2021gvinstightlycoupledgnssvisualinertial}. Existing evaluations, however, largely overlook how tracking performance degrades when visual or inertial sensing becomes unreliable for extended periods.

Beyond core visual–inertial estimation pipelines, a growing body of work has incorporated fault awareness into tightly coupled state estimation frameworks used for real-time tracking. Prior studies introduce explicit fault detection and isolation mechanisms, such as consistency checks and measurement gating, to identify abnormal sensor behavior and preserve estimator stability during anomalous operation \cite{fault-detection-of-resilient}. Complementary efforts investigate fault-tolerant multi-sensor fusion strategies that adapt estimator behavior through fault modeling, redundancy, or reweighting to improve robustness in urban and degraded sensing conditions \cite{aerospace10110923}. More recent approaches integrate learning-based components into classical filtering pipelines to correct estimation errors arising from complex visual failure modes \cite{Tabassum2023IntegratingGW}. Collectively, these methods are primarily designed to maintain estimator consistency or recover accuracy during fault events, rather than to analyze longer-term tracking behavior under sustained sensing degradation. \emph{Thus, there is a necessity for a pipeline that addresses accuracy under nominal conditions and addresses measurement noise, considering holistic sensor failure scenarios. Also, the pipeline should focus on detection rather than quantifying fault impact on accuracy, and provide explainability in terms of fault resilience.}

\section{Empirical Framework}
\label{sec:framework}

This section presents the empirical framework for evaluating sensor fault resilience in visual–inertial tracking systems for XR. The framework enables controlled study of tracking behavior under sustained sensing degradation while preserving realistic execution conditions.

\subsection{System Pipeline}

Our empirical study is conducted within a real-time visual–inertial tracking pipeline representative of inside-out XR systems, where pose estimation is driven by tightly coupled fusion of camera and inertial sensor measurements. The experimental setup preserves the sensing modalities, timing relationships, and execution structure commonly used in deployed XR tracking systems, allowing us to study tracking behavior under controlled sensing degradation without altering the estimator itself. We use ILLIXR \cite{illixr} solely as an implementation framework to support modular execution, synchronized sensor delivery, and reproducible experimentation; it is not the focus of our evaluation.

The tracking pipeline operates on synchronized visual and inertial inputs, including stereo camera and IMU streams. Camera frames provide visual observations at a moderate rate, while inertial measurements arrive at a significantly higher frequency and are used to propagate motion estimates between successive visual updates. This multi-rate sensing configuration reflects practical XR deployments and is central to how pose estimates evolve over time, particularly when one sensing modality becomes unreliable.

\begin{figure}[tb]
\centering
\includegraphics[width=\columnwidth]{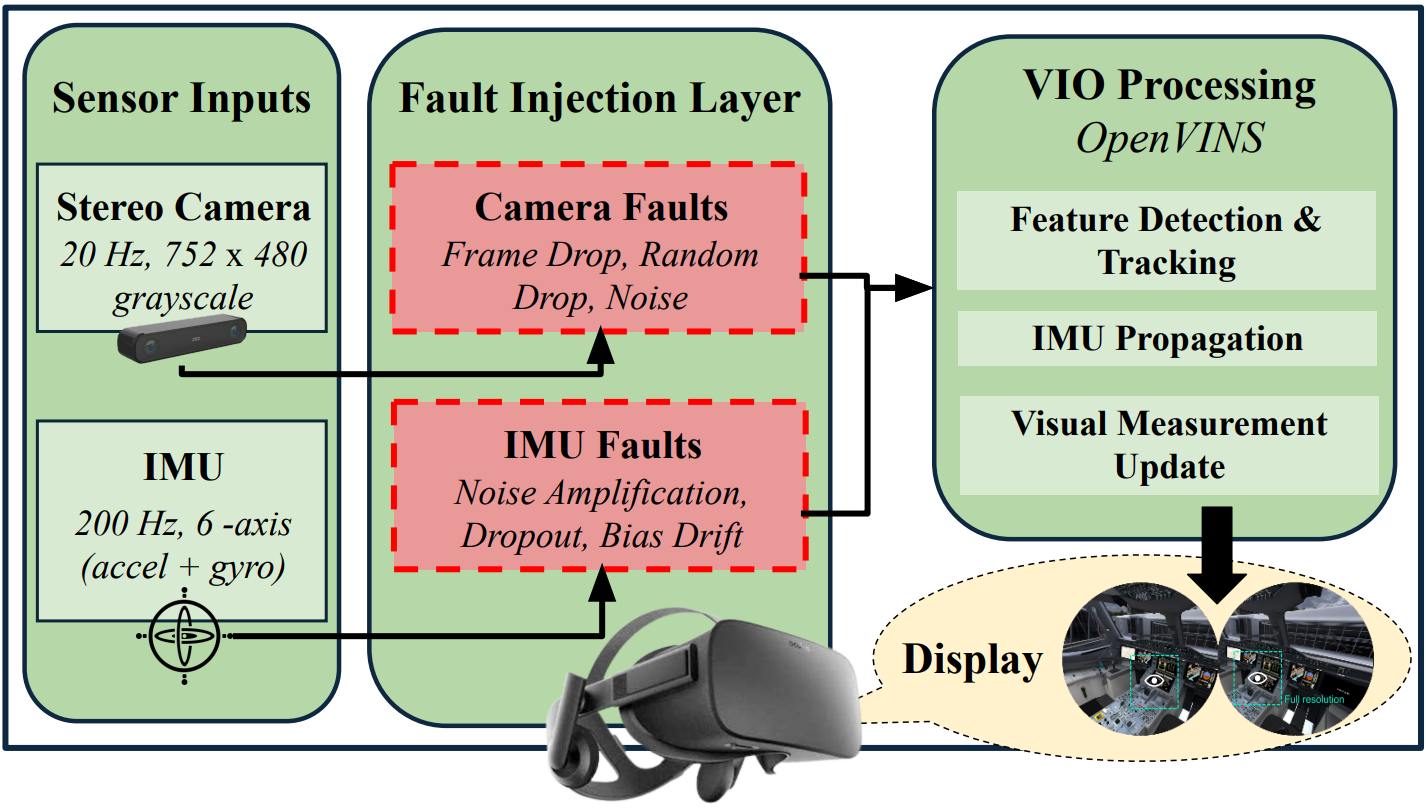}
\caption{System architecture of our empirical framework}
\label{fig:fig2}
\end{figure}

Visual and inertial measurements are fused using OpenVINS \cite{openvins}, which serves as the visual–inertial odometry implementation in the experiments. OpenVINS processes the incoming camera and IMU streams to produce a continuous pose estimate that would ordinarily be consumed by downstream XR components. This modular design allows us to isolate the effects of sensor degradation on tracking behavior. \cref{fig:fig2} provides an overview of the visual–inertial tracking pipeline and system context used in our experiments.

\subsection{Fault Injection Mechanism}
\label{sec:fault-injection}

To systematically study the resilience of visual–inertial tracking to sensor failures, we implement a fault-injection mechanism that operates directly on the camera and inertial data streams before they are consumed by the VIO estimator. This is shown as the \emph{Fault Injection Layer} in \cref{fig:fig2}. Faults are injected after raw sensor measurements are acquired but prior to visual–inertial processing, allowing us to emulate real-world sensing degradation while maintaining full experimental control. By intervening at the sensor interface, we precisely control which data the estimator receives while leaving the tracking pipeline itself unmodified.

Each fault scenario is defined using four parameters that together characterize the failure condition. The \emph{fault type} specifies the class of degradation being simulated, such as \texttt{data dropout}, \texttt{noise corruption}, or \texttt{bias drift}. The \emph{start} parameter determines when the fault is activated, specified relative to the camera frame index or IMU sample index, enabling evaluation of timing sensitivity. The \emph{duration} parameter controls how long the fault persists, allowing exploration of both short disruptions and sustained degradation. The \emph{severity} parameter governs the intensity of the fault effect, such as the probability of data loss or the magnitude of signal corruption. This parameterization enables systematic exploration of how individual fault characteristics influence tracking behavior and controlled comparisons across sensing modalities.

We consider several classes of camera-related faults that reflect common sources of visual degradation in XR systems. \emph{Consecutive camera frame drop} simulates a complete loss of visual input over a contiguous interval, as may occur due to physical occlusion, sensor crashes, or severe communication failures. During the fault window, incoming camera frames are discarded, and no visual information is delivered to the estimator, forcing tracking to rely entirely on inertial propagation. When visual input resumes, the system must reconcile accumulated inertial drift with newly observed image features, which may hinder reinitializing features and correcting pose. To capture a wide range of conditions, we evaluate frame drop durations spanning brief interruptions to extended visual outages.

In contrast to contiguous dropouts, \emph{random camera frame drop} models intermittent visual disruptions that occur stochastically within a time window. This fault pattern reflects scenarios such as packet loss, computational overload, or unstable sensing conditions, where some frames are lost while others are successfully delivered. Each incoming frame is dropped with a configurable probability, resulting in irregular gaps in the visual stream. While partial visual information remains available, skipped frames increase apparent inter-frame motion, potentially exceeding the tolerance of feature-tracking algorithms and degrading visual constraints. By varying the drop probability, we examine how increasing irregularity in visual availability affects tracking stability.

\begin{figure*}[!t]
    \centering

    \begin{minipage}[t]{0.24\textwidth}
        \centering
        \includegraphics[width=\linewidth]{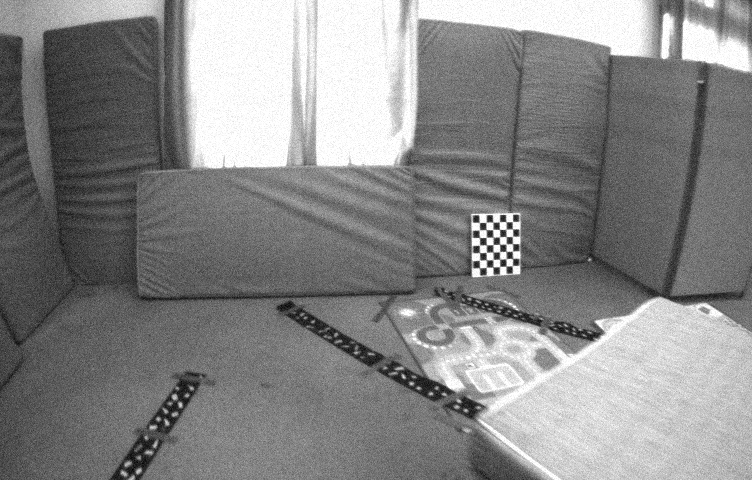}
        \subcaption{Left ($\sigma=10$)}
    \end{minipage}\hfill
    \begin{minipage}[t]{0.24\textwidth}
        \centering
        \includegraphics[width=\linewidth]{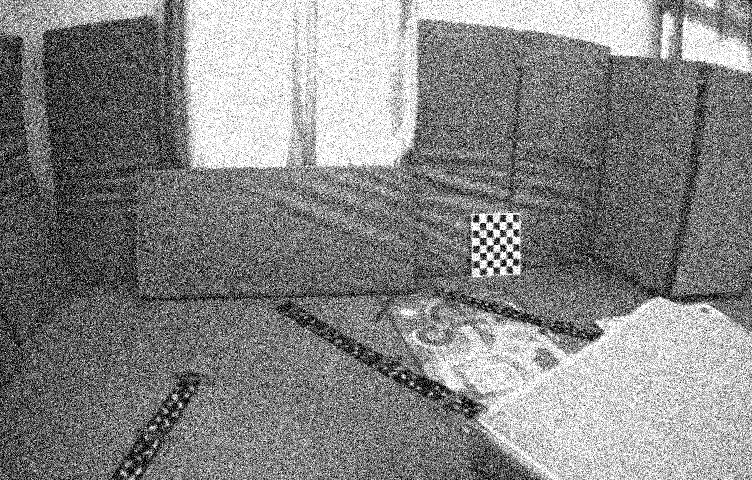}
        \subcaption{Left ($\sigma=50$)}
    \end{minipage}\hfill
    \begin{minipage}[t]{0.24\textwidth}
        \centering
        \includegraphics[width=\linewidth]{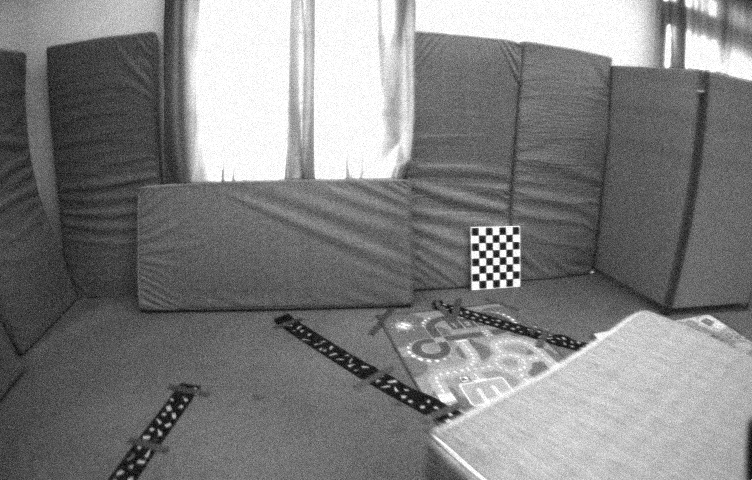}
        \subcaption{Right ($\sigma=10$)}
    \end{minipage}\hfill
    \begin{minipage}[t]{0.24\textwidth}
        \centering
        \includegraphics[width=\linewidth]{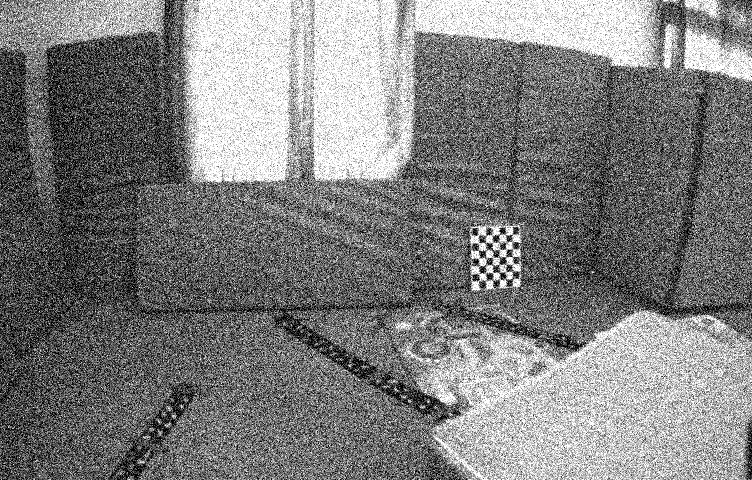}
        \subcaption{Right ($\sigma=50$)}
    \end{minipage}

    \caption{Stereo images with additive Gaussian noise. From left to right: left-camera view with low noise, left-camera view with high noise, right-camera view with low noise, and right-camera view with high noise.}
    \label{fig:cam_noise}
\end{figure*}

We also model degraded image quality through \emph{camera noise injection}. This fault represents visual corruption arising from environmental interference, optical contamination, low-light operation, or sensor aging. During the fault interval, additive Gaussian noise is applied independently to each pixel of the stereo images, with the noise magnitude controlled by the severity parameter. Unlike frame drop, noise injection preserves visual data but degrades its reliability, potentially reducing feature detection quality, increasing false matches, and weakening geometric constraints. This fault type allows us to study how gradual degradation of visual fidelity influences tracking behavior. \cref{fig:cam_noise} depicts how camera frames for both the left eye and the right eye cameras get affected by noise addition.

For inertial sensing, we consider several fault modes that affect the reliability of accelerometer and gyroscope measurements. \emph{IMU noise amplification} models increased measurement uncertainty caused by vibration, electromagnetic interference, or thermal effects. During the fault window, inertial measurements are scaled by a configurable factor to simulate conditions in which sensor outputs become unreliable without completely disappearing. Because inertial data directly drives motion propagation between camera updates, such corruption can rapidly distort predicted motion and challenge subsequent visual correction.

We also evaluate \emph{IMU dropout}, which represents the complete loss of inertial measurements due to hardware or communication failures. In this mode, no accelerometer or gyroscope data are delivered during the fault interval, forcing the system to rely solely on visual observations for pose estimation. This removes high-frequency motion priors, reducing robustness to fast motion and visual degradation, particularly during aggressive XR user movement. We evaluate short, yet complete, inertial outages that are representative of realistic transient failures.

Finally, we model \emph{IMU bias drift}, which represents gradual and persistent calibration errors that accumulate over time. Bias drift is particularly challenging because it introduces systematic error without producing obvious discontinuities in the sensor signal. In our implementation, a growing offset is added to gyroscope measurements during the fault window, causing the orientation error to increase progressively. This error propagates into velocity and position estimates due to gravity misalignment, making bias drift one of the most damaging fault modes in visual–inertial tracking.

Fault injection is controlled by a configuration file that specifies all fault parameters for camera and inertial sensors independently, enabling reproducible experiments without modifying the code. During execution, all fault events are logged with precise timestamps and frame or sample indices, allowing verification of fault activation and accurate alignment with pose estimates during analysis. The injection mechanism introduces negligible computational overhead, ensuring that observed effects arise from sensor degradation rather than from altered execution timing. This design enables controlled, repeatable evaluation of a wide range of sensor fault scenarios within a realistic visual–inertial XR tracking pipeline.

\subsection{Dataset and Reference Trajectory}

Our evaluation uses the \texttt{Vicon Room 1} sequence from the \emph{EuRoC MAV dataset} \cite{euroc}, shown in \cref{fig:cam_noise}. This dataset provides synchronized stereo camera and inertial measurements together with high-accuracy reference pose trajectories obtained from a motion capture system. The Vicon Room environment features structured indoor geometry and controlled lighting conditions representative of industrial and enterprise XR deployment settings, making it suitable for studying the impact of sensor faults on visual–inertial tracking.

Rather than directly comparing fault-injected trajectories against the reference poses, we adopt a baseline-relative evaluation methodology. A fault-free baseline trajectory is first generated by running the visual–inertial tracking pipeline on unmodified sensor data using identical initialization and estimator parameters. All fault-injected runs are then compared against this baseline trajectory, ensuring that observed deviations arise from injected sensor faults rather than from nominal estimation error or dataset-specific characteristics.


\subsection{Evaluation Metrics}

\label{sec:evaluation-metrics}

Our evaluation quantifies the impact of sensor faults by comparing fault-injected trajectories against a fault-free baseline using standard trajectory error metrics.
Each experiment produces a sequence of estimated six-degree-of-freedom poses, and deviations from the baseline trajectory are measured using both global and local error formulations.

\subsubsection{Absolute Trajectory Error (ATE)}
Absolute Trajectory Error (ATE) captures the global deviation between a fault-injected trajectory and the baseline trajectory. Given estimated poses $\mathbf{X}_i \in SE(3)$, where $ SE(3) $ denotes the space of rigid-body transformations in three dimensions, and corresponding baseline poses $\mathbf{X}_i^{b}$, the two trajectories are first aligned using a rigid-body transformation $\mathbf{A} \in SE(3)$. The pose error at time step $i$ is defined as

\begin{equation}
\mathbf{E}_i = \mathbf{X}_i^{b^{-1}} \, \mathbf{A} \, \mathbf{X}_i .
\end{equation}

The translational component of this error is used to compute the mean ATE:

\begin{equation}
\mathrm{ATE}_{\mathrm{mean}}
=
\frac{1}{N}
\sum_{i=1}^{N}
\left\lVert \operatorname{pos}(\mathbf{E}_i) \right\rVert
\end{equation}

where $\text{pos}(\cdot)$ extracts the three-dimensional position vector and $N$ is the total number of poses. ATE reflects the cumulative effect of tracking error over the full trajectory.

\subsubsection{Relative Pose Error (RPE)}
To assess local motion accuracy, we compute Relative Pose Error (RPE), which measures deviations in relative motion over a fixed interval $\Delta$. For each valid pose pair, the relative pose error is defined as

\begin{equation}
\mathbf{R}_i
=
\left(
\mathbf{X}_i^{b^{-1}}
\mathbf{X}_{i+\Delta}^{b}
\right)^{-1}
\left(
\mathbf{X}_i^{-1}
\mathbf{X}_{i+\Delta}
\right)
\end{equation}

The translational RPE is summarized using the mean error:

\begin{equation}
\mathrm{RPE}_{\mathrm{mean}}(\Delta)
=
\frac{1}{K}
\sum_{i=1}^{K}
\left\lVert \operatorname{pos}(\mathbf{R}_i) \right\rVert
\end{equation}

where $K$ denotes the number of valid pose pairs. RPE characterizes short-term drift behavior and complements the global ATE metric.

In addition to aggregate metrics, we examine the temporal evolution of position and orientation errors, aligned with the fault injection interval, to understand how errors emerge, evolve, and persist over time. To account for run-to-run variability, experiments are repeated when stochastic effects are present; representative or median results are reported accordingly, with all runs using an identical configuration and initialization.

\subsection{Experimental Configuration and Implementational Details}
\label{sec:experimental-config}

Our evaluation considers a fixed set of experimental configurations that instantiate the fault models described in \cref{sec:fault-injection} across representative choices of timing, duration, and severity. The full set of configurations is summarized in \cref{tab:exp_config}, which lists the fault categories, parameter ranges, and execution settings used in our experiments.

All configurations are evaluated using the same visual–inertial tracking pipeline, dataset segment, initialization procedure, and evaluation metrics, enabling direct comparison of fault impact across conditions. Where stochastic fault behavior is involved, multiple runs are executed, and representative statistics are reported as described in \cref{sec:evaluation-metrics}. This structured configuration design supports systematic comparison across fault types while keeping the experimental methodology compact and reproducible.

\begin{table*}[!t]                                                                                 
  \centering                                                                                         
  \caption{Summary of experimental fault configurations evaluated in this study.}                    
  \label{tab:exp_config}                                                                             
  \begin{tabular}{l l l l l}                                                                         
  \hline                                                                                             
  \textbf{Category} & \textbf{Fault Type} & \textbf{Variations} & \textbf{Key Parameters} &          
  \textbf{Research Focus} \\                                                                         
  \hline                                                                                             
  Timing    & Camera drop & Start / Mid / End & 50 frames at 100 / 600 / 1000 & Effect of fault      
  timing \\                                                                                          
  Timing    & IMU noise   & Start / Mid / End & 200 samples at 1000 / 6000 / 10000, 5$\times$ &      
  Effect of fault timing \\                                                                          
  Duration  & Camera drop & 6 levels          & 5 / 10 / 20 / 50 / 100 / 200 frames & Visual blackout
   tolerance \\                                                                                      
  Duration  & IMU dropout & 5 levels          & 50 / 100 / 200 / 400 / 800 samples & Inertial outage 
  tolerance \\                                                                                       
  Random    & Camera drop & 4 levels          & 10\% / 20\% / 30\% / 50\% drop rate & Intermittent   
  vs. continuous loss \\                                                                             
  Severity  & IMU noise   & 3 levels          & 2$\times$ / 5$\times$ / 10$\times$ amplification &   
  Inertial noise tolerance \\                                                                        
  Drift     & IMU bias    & 3 levels          & 0.01 / 0.05 / 0.1 rad/s/s & Sensitivity to           
  calibration drift \\                                                                               
  Noise     & Camera      & 3 levels          & Gaussian noise std. dev. 10 / 30 / 50 & Visual       
  degradation tolerance \\                                                                           
  Combined  & Camera + IMU & Single case      & Camera drop + IMU noise & Compound fault interaction 
  \\                                                                                                 
  \hline                                                                                             
  \end{tabular}                                                                                      
  \end{table*} 

All experiments were performed on a machine with a 24-core CPU, 32 GB of RAM, and an NVIDIA RTX 2000 Ada GPU, running Ubuntu 22.04. The ILLIXR runtime was used to emulate the XR headset environment during execution.

\section{Experimental Analysis}

\label{sec:expertimental}

This section presents an empirical analysis of how sustained sensor faults affect visual–inertial tracking behavior in XR systems. We report results across the fault configurations summarized in \cref{sec:experimental-config}, focusing on trajectory error magnitude, temporal evolution, and differences between camera and inertial degradation.

\subsection{Effect of Fault Injection Timing}

We evaluate how the timing of sensor faults along the trajectory affects visual–inertial tracking accuracy. Identical fault conditions are injected at three points in the trajectory, early (t = 5 s), mid (t = 30 s), and late (t = 50 s), while fault type, duration, and severity are held constant.

\subsubsection{Camera Fault Timing}

\begin{figure*}[!t]
    \centering
    \setlength{\tabcolsep}{0pt} 
    \renewcommand{\arraystretch}{0} 

    \begin{tabular}{@{}ccc@{}}
        \begin{subfigure}[t]{0.362\textwidth}
            \centering
            \includegraphics[width=\linewidth]{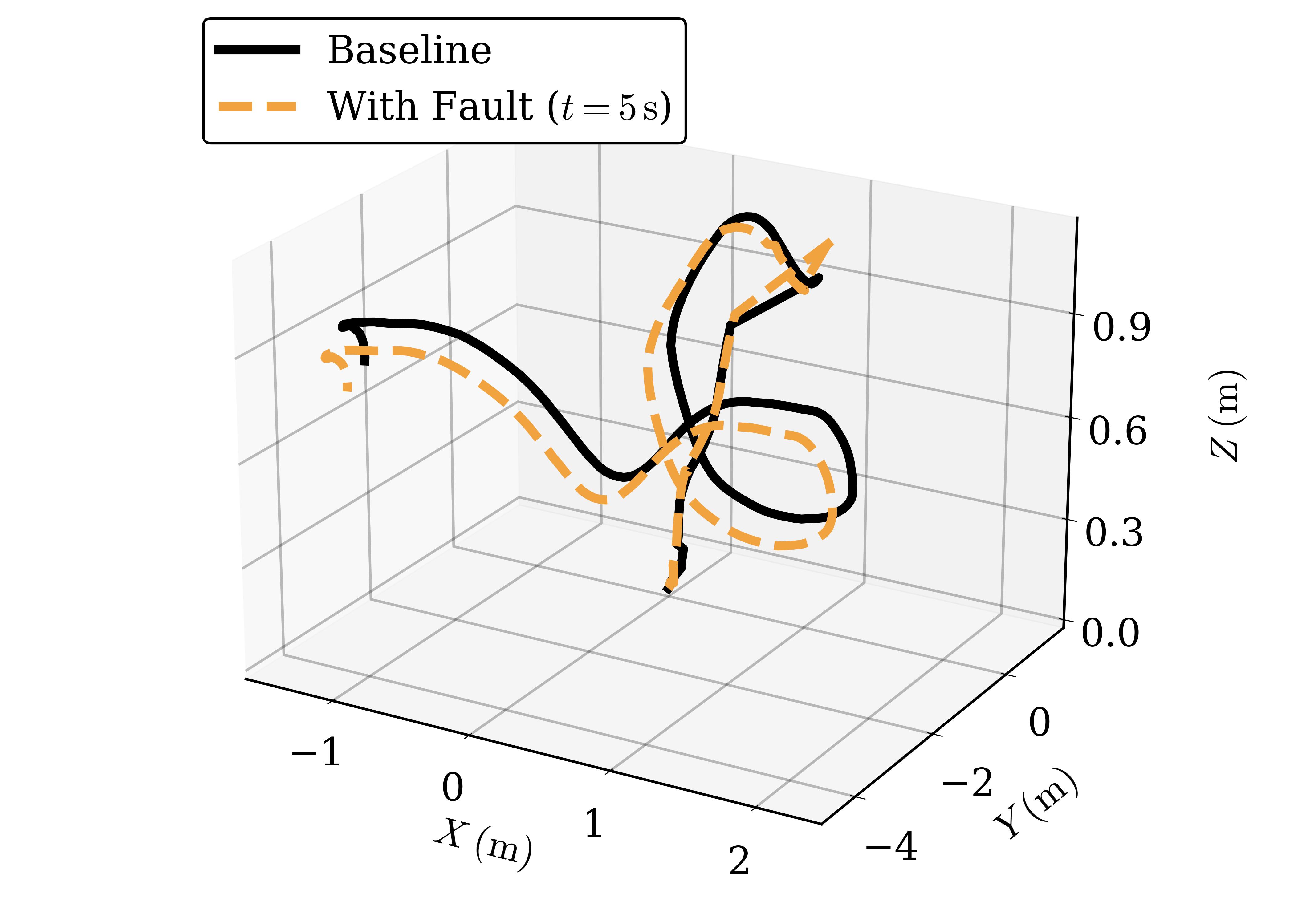}
            \vspace{-2mm}
            \caption{Start ($t=5$s), Mean ATE: 12.46 cm}
            \label{fig:cam-fault-traj-start}
        \end{subfigure}
        &
        \begin{subfigure}[t]{0.362\textwidth}
            \centering
            \includegraphics[width=\linewidth]{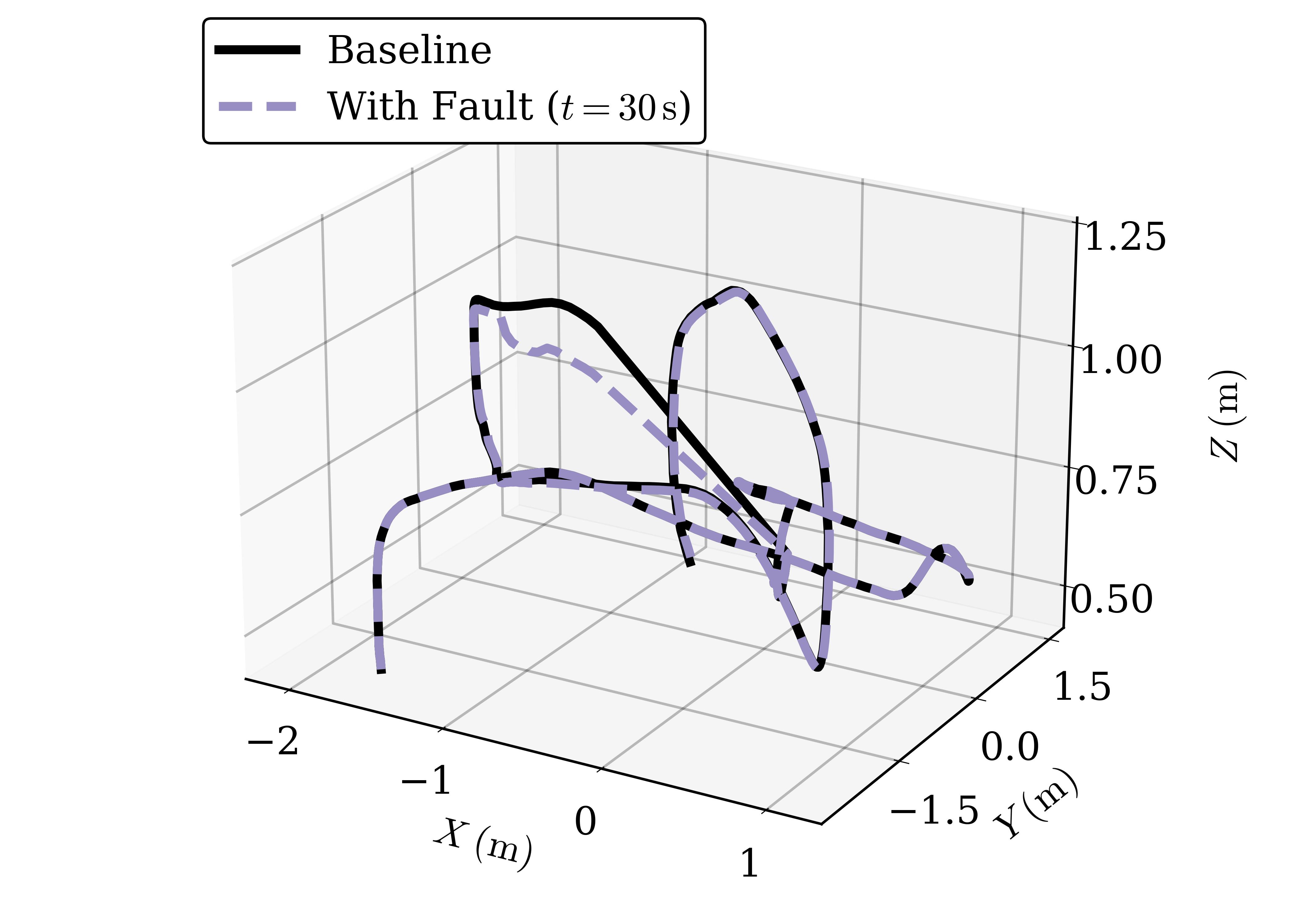}
            \vspace{-2mm}
            \caption{Middle ($t=30$s), Mean ATE: 2.56 cm}
            \label{fig:cam-fault-traj-middle}
        \end{subfigure}
        &
        \begin{subfigure}[t]{0.362\textwidth}
            \centering
            \includegraphics[width=\linewidth]{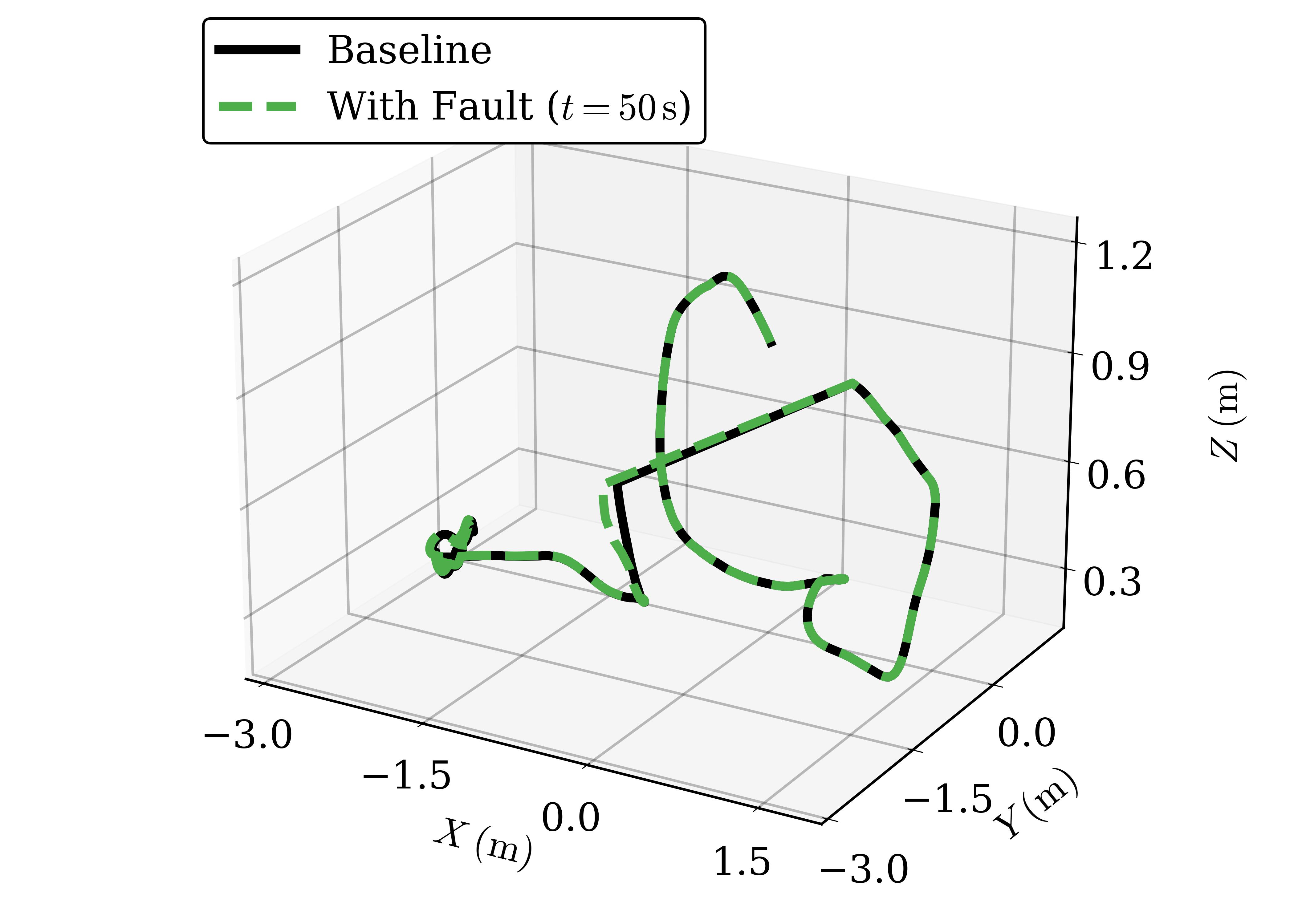}
            \vspace{-2mm}
            \caption{End ($t=50$s), Mean ATE: 0.59 cm}
            \label{fig:cam-fault-traj-end}
        \end{subfigure}
    \end{tabular}

    \vspace{-2mm}
    \caption{Three-dimensional trajectory segments under camera frame drop faults injected at different times along the trajectory. Each panel shows a local view centered on the fault injection region, comparing the fault-free baseline trajectory (black) with the fault-affected trajectory (colored). Red markers indicate the fault injection locations.}
    \vspace{-2mm}
    \label{fig:cam-fault-traj}
\end{figure*}


\cref{fig:cam-fault-traj} compares three-dimensional trajectories under camera frame-drop faults injected at different times, where the fault-free baseline trajectory is shown in black and the fault-affected trajectory is shown in color; fault locations correspond to the points at which the fault-affected trajectory begins to deviate from the baseline. When camera frames are dropped early in the trajectory ($t = 5\,\mathrm{s}$), the estimated path exhibits a pronounced deviation, resulting in a mean translational ATE of $12.46\,\mathrm{cm}$. Faults injected at the midpoint of the trajectory ($t = 30\,\mathrm{s}$) lead to a smaller deviation, with a mean translational ATE of $2.56\,\mathrm{cm}$, while faults injected near the end of the trajectory ($t = 50\,\mathrm{s}$) have minimal impact, yielding a mean translational ATE of $0.59\,\mathrm{cm}$. Across all cases, camera-induced errors remain bounded within the centimeter scale, with earlier faults producing larger accumulated errors due to the longer remaining trajectory over which residual estimation errors propagate.

\subsubsection{IMU Fault Timing}

\begin{figure}[!t]
    \centering
    \includegraphics[width=0.9\columnwidth]{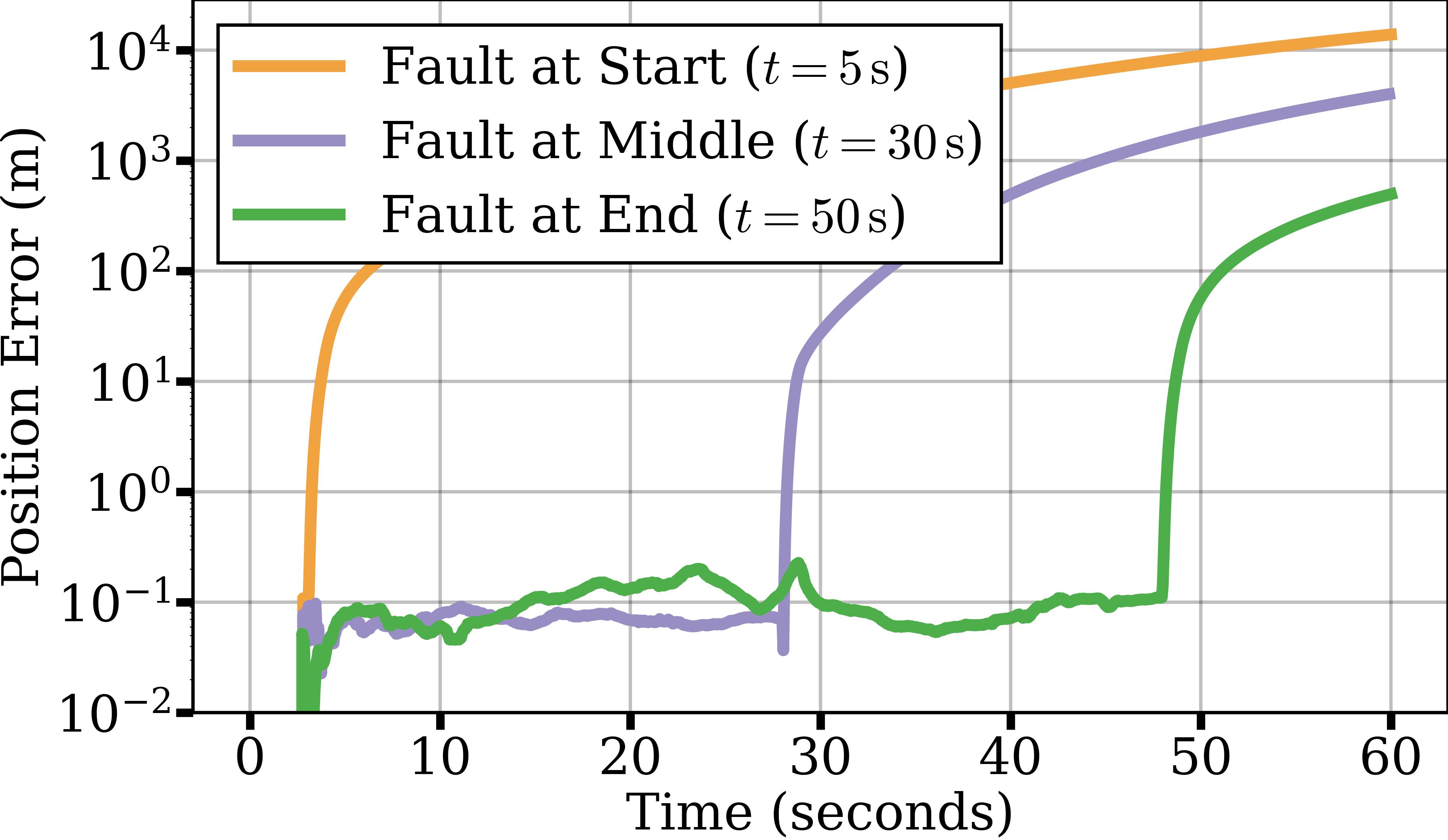}
    \caption{Temporal evolution of translational ATE under IMU noise faults injected at t = 5 s, 30 s, and 50 s. Error is shown on a logarithmic scale to capture the wide range of observed magnitudes across fault timing conditions.}
    \label{fig:imu-fault-time}
\end{figure}


\cref{fig:imu-fault-time} shows the temporal evolution of mean translational ATE under IMU noise faults injected at the same three trajectory points. In contrast to camera faults, IMU faults produce rapid and unbounded error growth, with position errors spanning multiple orders of magnitude; the small deviation observed prior to fault injection reflects inherent estimator variability under nominal conditions. Early IMU faults ($t = 5\,\mathrm{s}$) lead to continuous error growth over the entire trajectory, reaching a mean translational ATE of $3{,}982\,\mathrm{m}$ by the end of execution, while faults injected at the midpoint ($t = 30\,\mathrm{s}$) follow a similar trend but accumulate less error, resulting in a mean translational ATE of $686\,\mathrm{m}$. Even late faults ($t = 50\,\mathrm{s}$) produce substantial deviation, yielding a mean translational ATE of $46\,\mathrm{m}$ despite the limited remaining trajectory duration. Across all timing conditions, IMU faults induce errors that are three to four orders of magnitude larger than those caused by comparable camera faults, highlighting a pronounced asymmetry in fault impact across sensor modalities.

\subsection{Effect of Camera Blackout Duration}

We examine how the duration of consecutive camera frame loss affects visual–inertial tracking accuracy. Camera frames are dropped starting at t = 25 s, with blackout durations ranging from 0.25 s (5 frames) to 10 s (200 frames), while all other conditions remain fixed.

\begin{figure}[!t]
    \centering
    \includegraphics[width=\columnwidth]{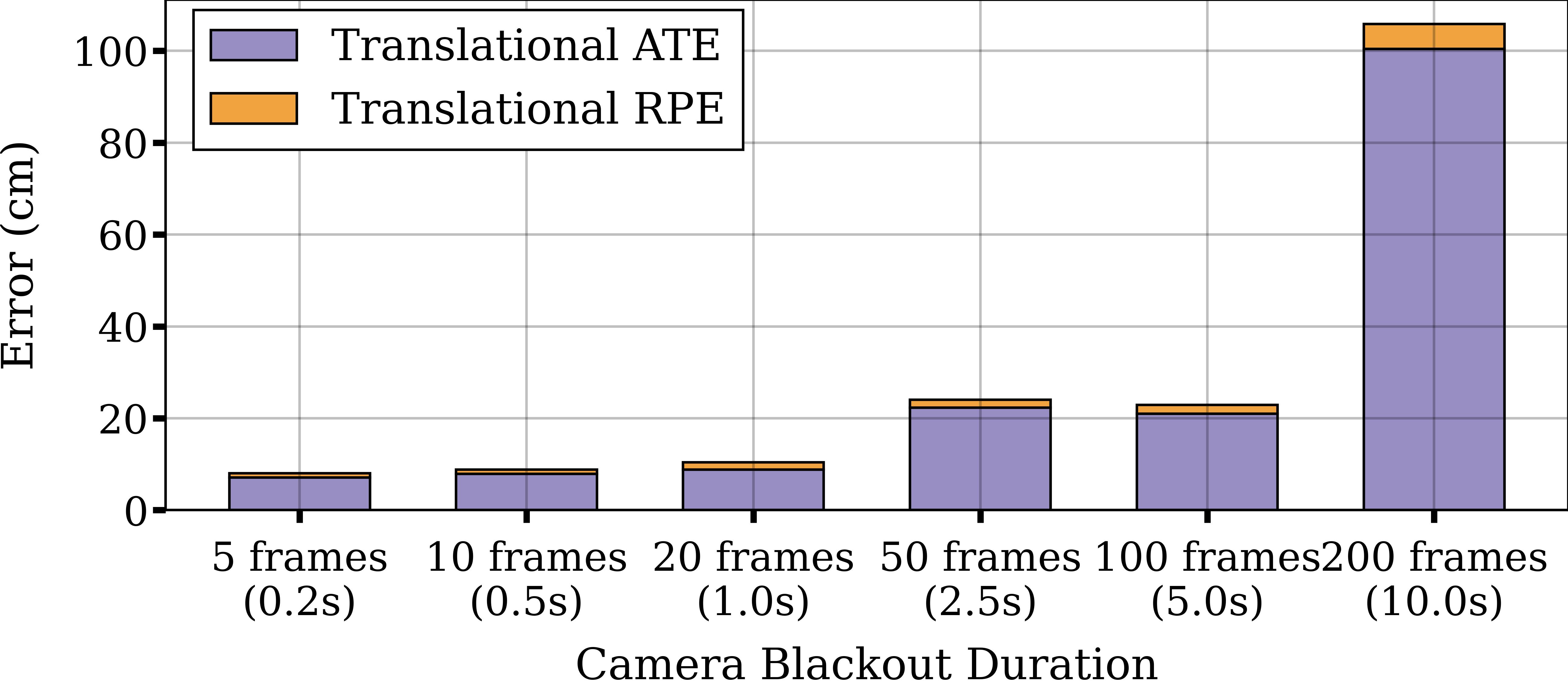}
    \caption{Mean translational ATE and RPE as a function of camera blackout duration. ATE increases with longer blackouts, reaching over 1 m at 10 s, while RPE remains comparatively low across all durations.}
    \label{fig:cam-blackout-duration}
\end{figure}

\cref{fig:cam-blackout-duration} reports the mean translational Abscauses catastrophic degradation Relative Pose Error (RPE) across the six blackout durations. Translational ATE increases with blackout duration, from 7.05 cm at 0.25 s to 100.42 cm at 10 s. Short blackouts up to 1.0 s result in limited cumulative error below 10 cm, while longer blackouts lead to substantially larger deviations, including 22.28 cm ATE at 2.5 s and over 1 m ATE at 10 s.

In contrast, translational RPE remains comparatively low across all conditions, ranging from 0.88 cm to 5.40 cm. Even during extended blackouts, RPE increases modestly relative to ATE, indicating that short-term relative-motion estimates remain stable after camera input resumes.

\subsection{Effect of IMU Dropout Duration}

\begin{figure*}[!t]
\centering
\begin{subfigure}[t]{0.49\textwidth}
  \centering
  \includegraphics[width=\linewidth]{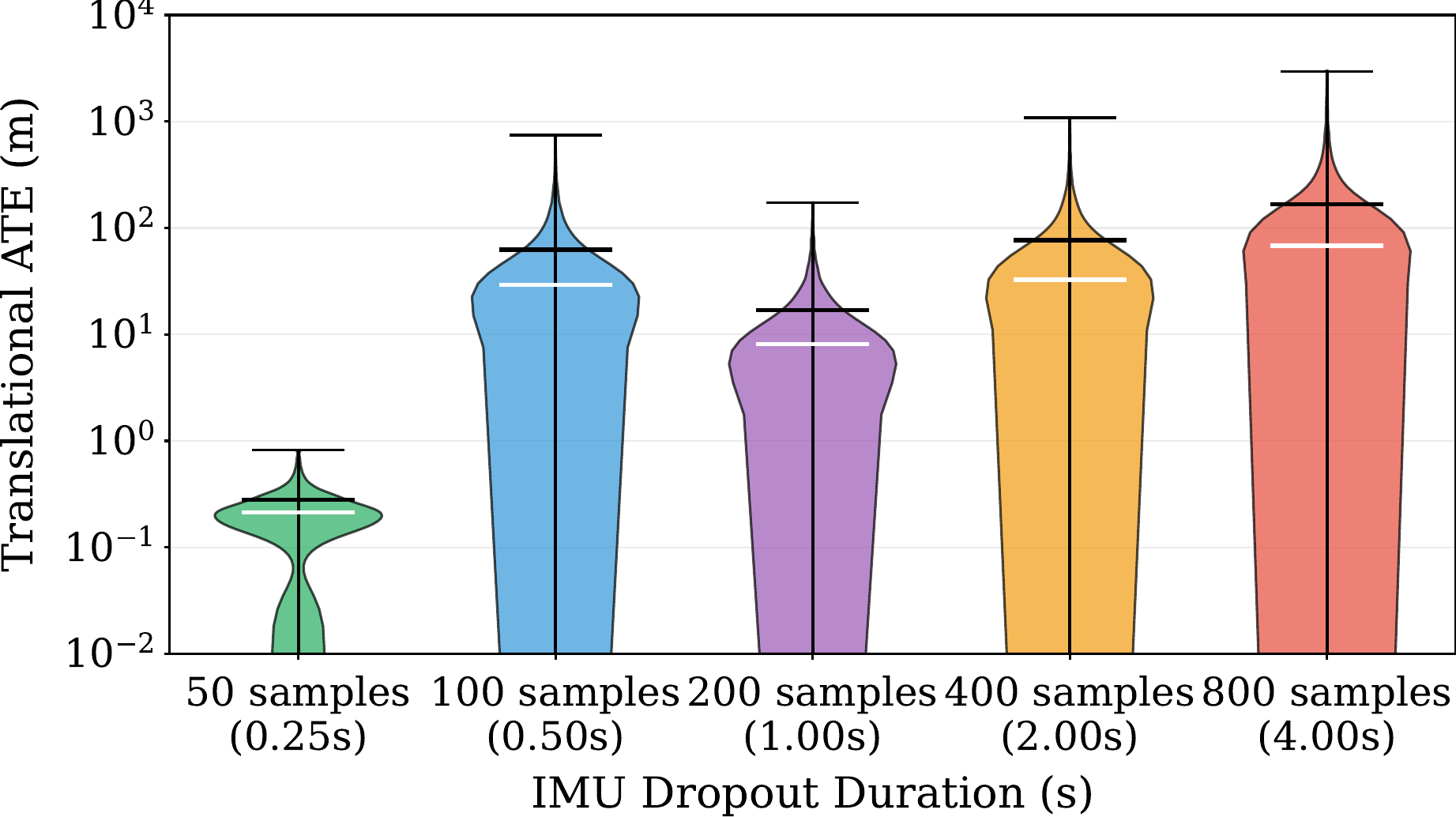}
  \caption{Mean Translational ATE across IMU dropout durations.}
\end{subfigure}\hfill
\begin{subfigure}[t]{0.49\textwidth}
  \centering
  \includegraphics[width=\linewidth]{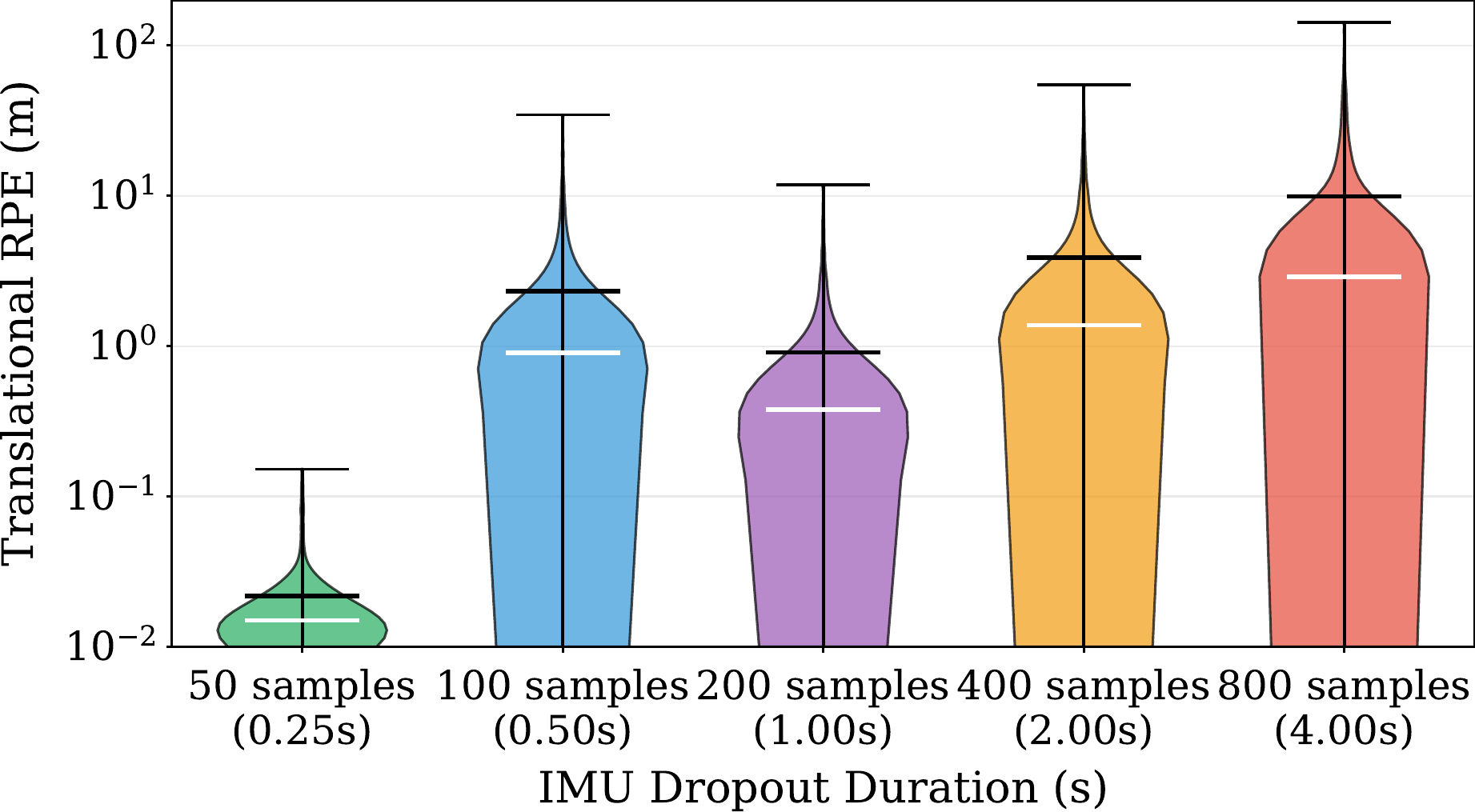}
  \caption{Mean Translational RPE across IMU dropout durations.}
\end{subfigure}

\vspace{-1mm}
\caption{Impact of IMU dropout duration on visual-inertial tracking accuracy.}
\label{fig:imu_dropout}
\end{figure*}

We evaluate the impact of consecutive IMU measurement dropouts of varying duration on visual-inertial tracking accuracy. IMU samples are dropped starting at $t = 25\,\mathrm{s}$, with dropout durations ranging from $0.25\,\mathrm{s}$ (50 samples) to $4\,\mathrm{s}$ (800 samples) at a 200~Hz IMU rate. \cref{fig:imu_dropout} summarizes the resulting mean translational ATE and RPE distributions across dropout durations, respectively. Very short outages of $0.25\,\mathrm{s}$ result in minimal degradation, with a mean ATE of $0.18\,\mathrm{m}$ and RPE of $0.03\,\mathrm{m}$, indicating that the estimator can briefly rely on visual measurements alone. In contrast, longer IMU dropouts cause severe tracking failure, with errors increasing by several orders of magnitude. A $0.5\,\mathrm{s}$ dropout produces $150.53\,\mathrm{m}$ ATE, while a $4\,\mathrm{s}$ dropout results in $721.07\,\mathrm{m}$ ATE, with corresponding RPE values reaching $31.51\,\mathrm{m}$.

A non-monotonic trend is observed at intermediate durations: a $1.0\,\mathrm{s}$ dropout produces lower error ($58.59\,\mathrm{m}$ ATE and $2.55\,\mathrm{m}$ RPE) than the $0.5\,\mathrm{s}$ case. This behavior is consistently reflected in both ATE and RPE distributions, indicating that it arises from estimator dynamics rather than cumulative error artifacts. 
Overall, these results demonstrate that IMU dropout causes catastrophic degradation relative to camera outages, underscoring the critical role of inertial sensing in maintaining VIO stability.

\subsection{Effect of Random Camera Frame Drops}

\begin{figure}[!t]
    \centering
    \includegraphics[width=0.7\columnwidth]{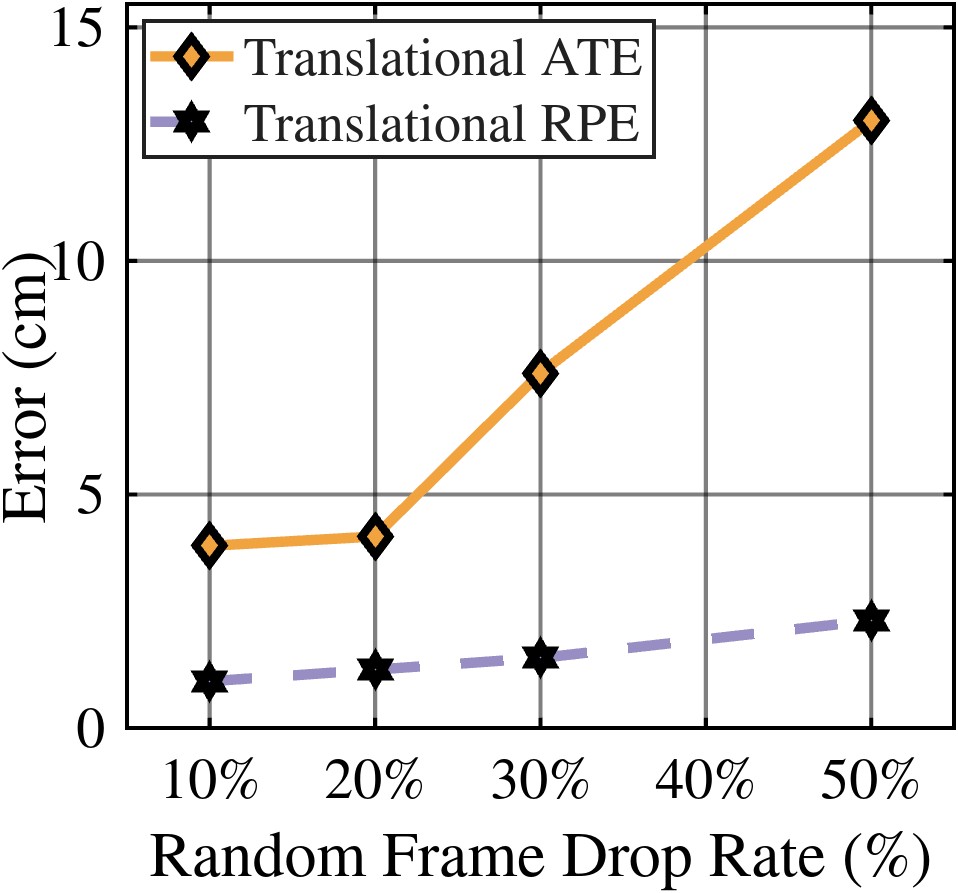}
    \caption{Mean Translational ATE and RPE under random camera frame drops at rates of 10\%, 20\%, 30\%, and 50\%. Both metrics increase monotonically with higher drop rates while remaining bounded across all conditions.}
    \label{fig:random-cam-frame-drop}
\end{figure}


\begin{figure}[!t]
    \centering
    \includegraphics[width=0.7\columnwidth]{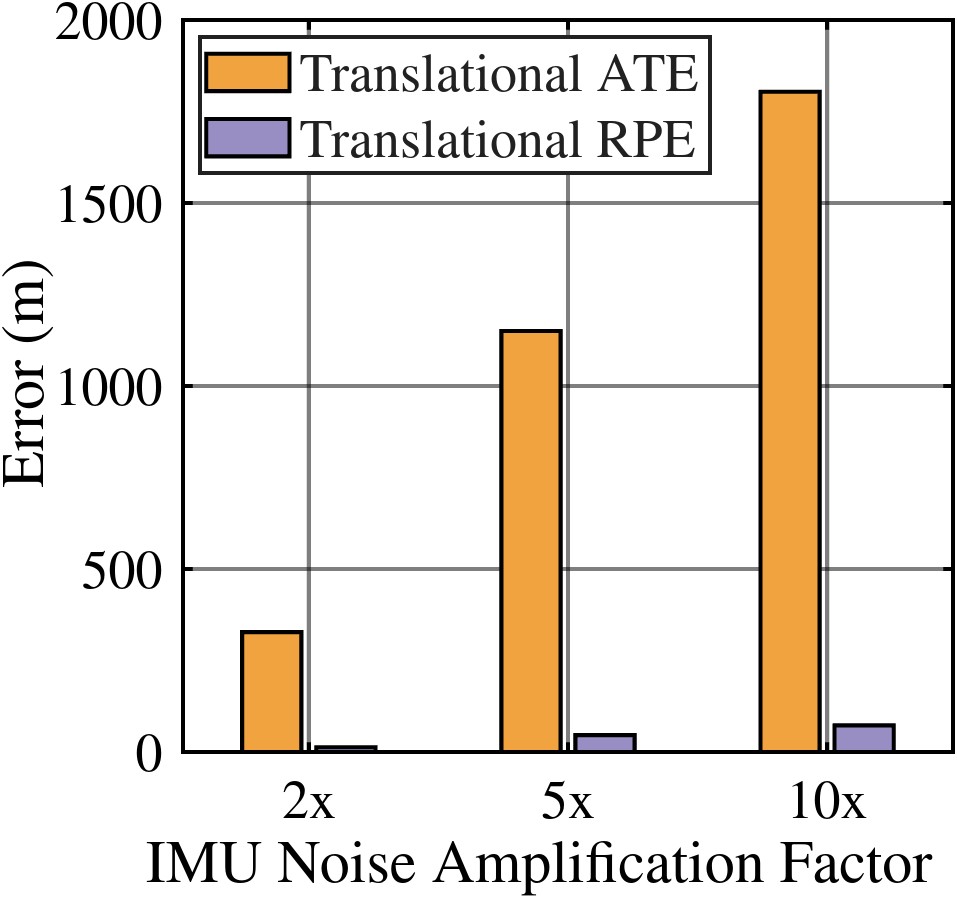}
    \caption{Mean translational ATE and RPE under IMU noise amplification. Results are shown for noise scaling factors of $2\times$, $5\times$, and $10\times$. Both metrics increase rapidly with noise level, reaching kilometer-scale errors at higher amplification.}
    \label{fig:imu_noise_amplification}
\end{figure}

We evaluate the impact of stochastic camera frame loss on VIO accuracy by randomly dropping frames throughout the trajectory with probabilities of $10\%$, $20\%$, $30\%$, and $50\%$. Unlike consecutive blackouts, random drops distribute visual outages temporally, avoiding extended periods without visual updates. \cref{fig:random-cam-frame-drop} reports mean translational ATE and mean translational RPE under these conditions. Both metrics increase monotonically with drop rate, indicating progressively degraded performance as camera reliability decreases. Mean translational ATE increases from $3.90\,\mathrm{cm}$ at a $10\%$ drop rate to $12.99\,\mathrm{cm}$ at $50\%$, while mean translational RPE rises from $0.97\,\mathrm{cm}$ to $2.28\,\mathrm{cm}$ over the same range.

Despite substantial frame loss, errors remain bounded, with mean translational ATE below $15\,\mathrm{cm}$ even at a $50\%$ random drop rate, indicating strong robustness to temporally distributed visual outages. In contrast to consecutive camera blackouts, random drops result in significantly lower error despite a larger total number of lost frames. For example, a $50\%$ random drop condition (approximately $600$ dropped frames) yields a mean translational ATE of $12.99\,\mathrm{cm}$, whereas a $10\,\mathrm{s}$ consecutive blackout (approximately $200$ dropped frames) produces a mean translational ATE exceeding $100\,\mathrm{cm}$. This contrast indicates that the temporal distribution of frame loss strongly influences fault impact.

\subsection{Effect of IMU Noise Amplification}

\begin{figure*}[!t]
    \centering
    \begin{subfigure}[t]{0.49\textwidth}
        \centering
        \includegraphics[width=\linewidth]{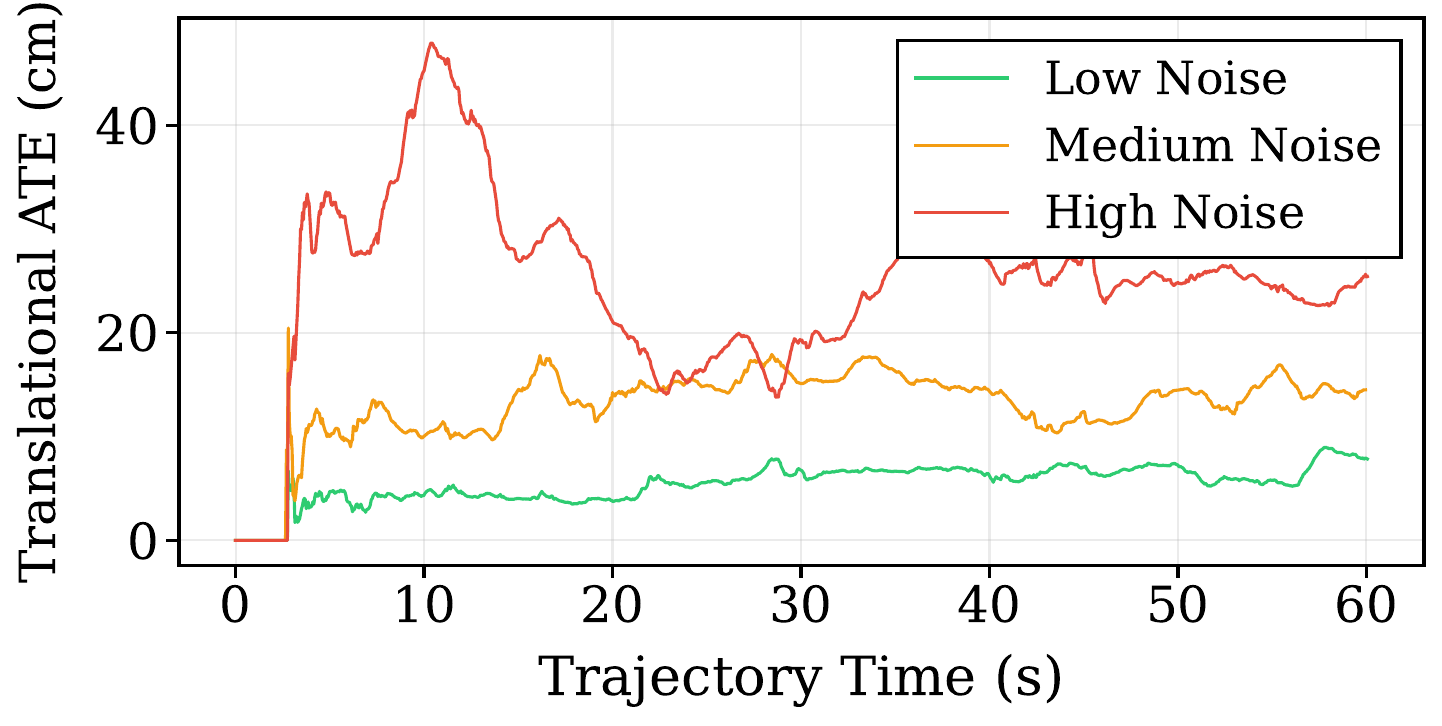}
        \caption{Mean Translational ATE under camera image noise. Error increases with noise intensity while remaining bounded.}
        \label{fig:cam_noise_ate}
    \end{subfigure}\hfill
    \begin{subfigure}[t]{0.49\textwidth}
        \centering
        \includegraphics[width=\linewidth]{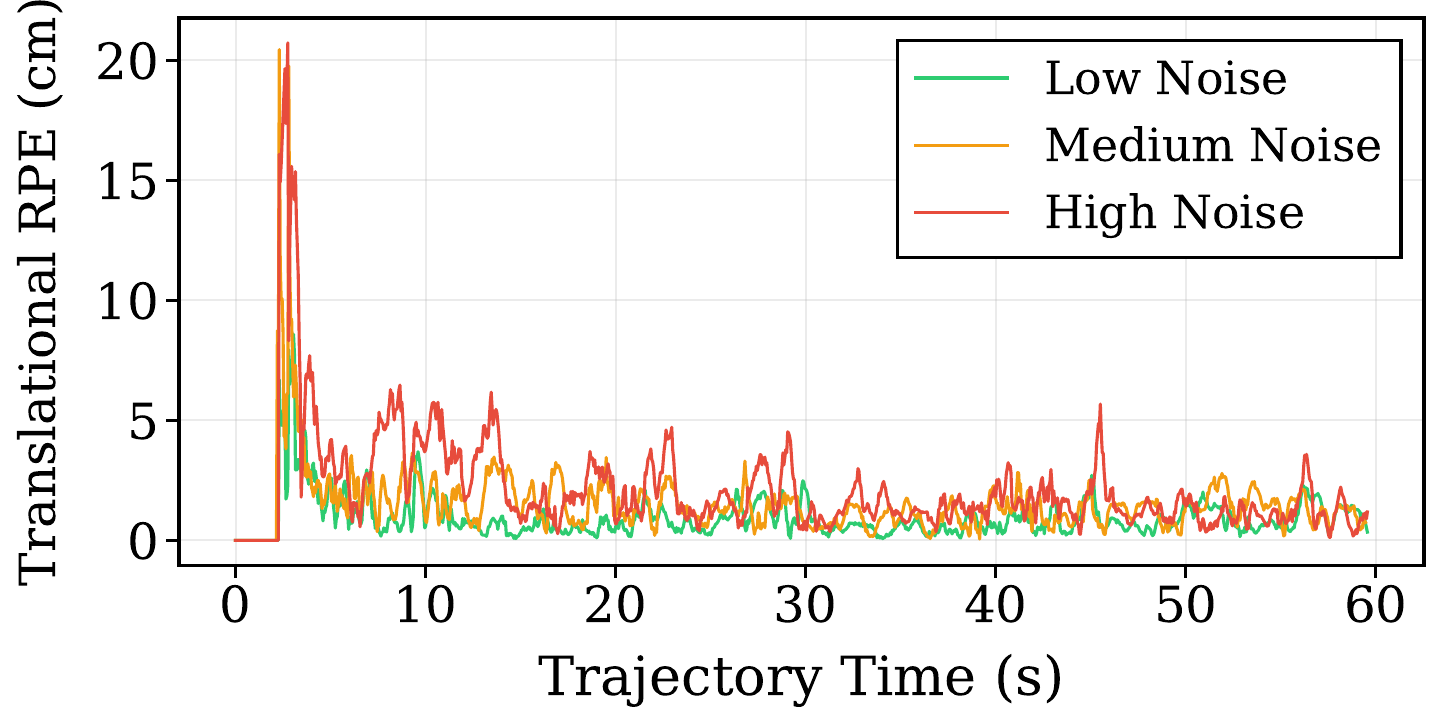}
        \caption{Mean Translational RPE under camera image noise. Local tracking error degrades gradually with noise level.}
        \label{fig:cam_noise_rpe}
    \end{subfigure}
    \vspace{-1mm}
    \caption{Impact of camera image noise on VIO accuracy. Both cumulative (ATE) and local (RPE) errors increase monotonically with noise intensity while remaining within the centimeter range.}
    \label{fig:camera-noise}
\end{figure*}



We evaluate VIO robustness to degraded IMU signal quality by artificially amplifying sensor noise by factors of $2\times$, $5\times$, and $10\times$ beginning at $t = 25\,\mathrm{s}$, simulating conditions such as sensor calibration drift, electromagnetic interference, or hardware degradation affecting accelerometer and gyroscope measurements. \cref{fig:imu_noise_amplification} presents a grouped bar chart comparing mean translational ATE and mean translational RPE across the three noise amplification levels, showing that both metrics increase sharply with higher noise. At $2\times$ noise amplification, the system already exhibits severe degradation, with a mean translational ATE of $327.49\,\mathrm{m}$ and a mean translational RPE of $13.04\,\mathrm{m}$, indicating strong sensitivity of VIO performance to inertial measurement quality. The elevated mean translational RPE demonstrates that IMU noise affects not only cumulative drift but also local tracking accuracy by corrupting the motion model used for inter-frame feature prediction. Increasing the noise to $5\times$ results in a mean translational ATE of $1150.38\,\mathrm{m}$ and a mean translational RPE of $46.01\,\mathrm{m}$, while $10\times$ amplification further escalates the errors to a mean translational ATE of $1804.11\,\mathrm{m}$ and a mean translational RPE of $72.66\,\mathrm{m}$ over a $60\,\mathrm{s}$ trajectory within a $3\,\mathrm{m} \times 3\,\mathrm{m}$ workspace. Overall, increasing IMU noise levels lead to progressively larger errors in both global and local trajectory estimates.

\subsection{Effect of Camera Image Noise}

We evaluate VIO robustness to degraded image quality by injecting additive Gaussian noise into camera frames throughout the trajectory. Three noise levels are considered (low, medium, and high), representing increasing visual degradation caused by sensor noise or adverse imaging conditions.

\cref{fig:cam_noise_ate} and \cref{fig:cam_noise_rpe} report the temporal evolution of translational ATE and translational RPE under the three noise settings, respectively. Both metrics increase monotonically with noise intensity, but remain bounded within the centimeter range across all conditions. Mean translational ATE increases from 5.45 cm under low noise to 13.02 cm under medium noise and 24.78 cm under high noise. Correspondingly, translational RPE increases from 0.95 cm to 1.51 cm and 2.15 cm, indicating progressively degraded local motion estimation with increasing image corruption.

While camera noise introduces larger errors than random camera frame drops at comparable visual degradation levels (e.g., 24.78 cm ATE under high noise versus 12.99 cm ATE at 50\% random frame loss), the resulting errors remain orders of magnitude smaller than those caused by IMU faults. Even at the highest noise level, camera-induced errors remain below 25 cm, whereas under IMU noise amplification, they can reach hundreds of meters. These results show that visual degradation leads to gradual, bounded error growth, in contrast to the catastrophic divergence observed under inertial-sensing faults.

\subsection{Effect of IMU Bias Drift}

\begin{figure}[!t]
    \centering
    \begin{subfigure}[t]{0.48\columnwidth}
        \centering
        \includegraphics[width=\linewidth]{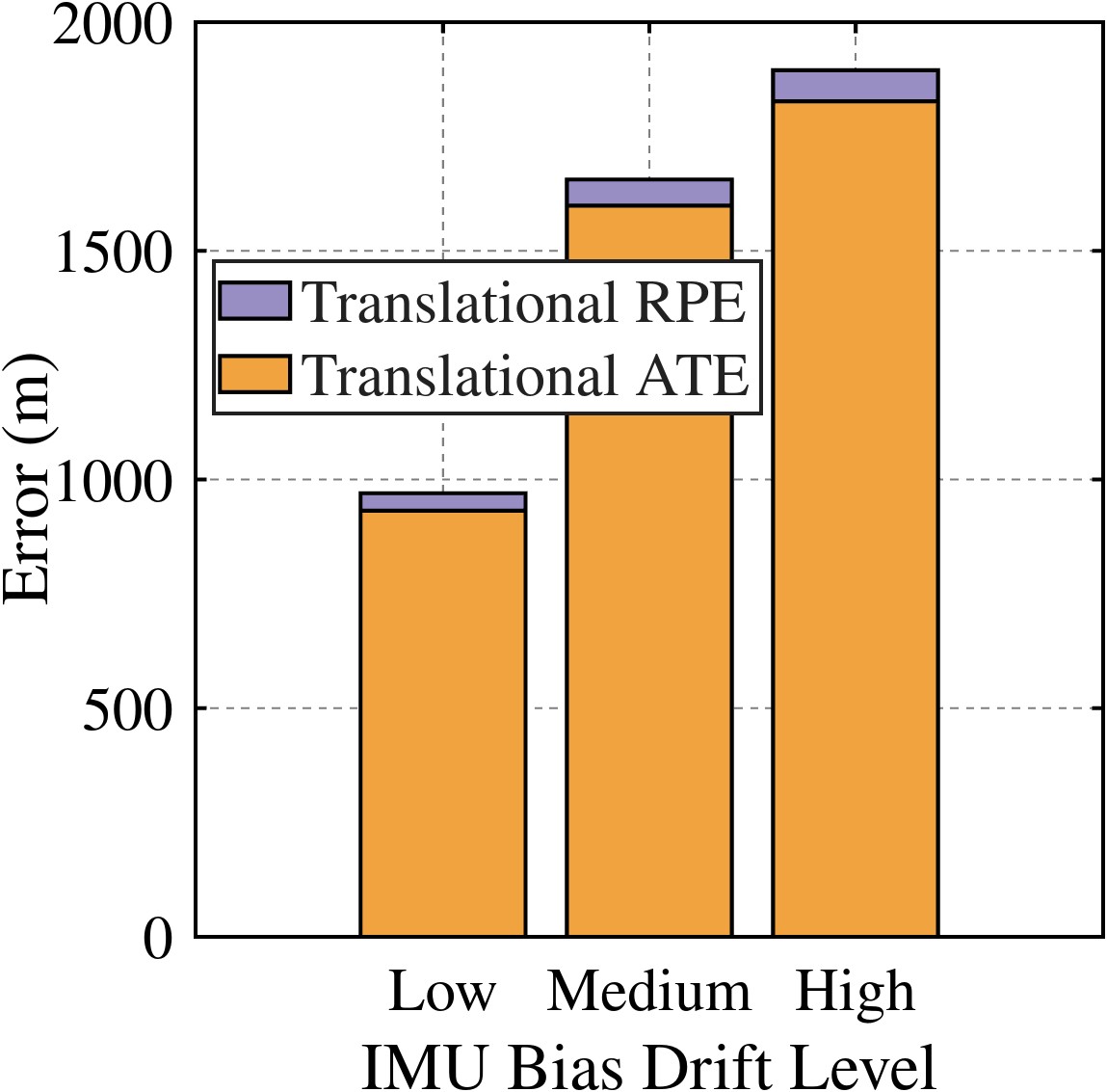}
        \caption{Mean Translational ATE and RPE under IMU bias drift.}
        \label{fig:imu_bias_ate_rpe}
    \end{subfigure}\hfill
    \begin{subfigure}[t]{0.48\columnwidth}
        \centering
        \includegraphics[width=\linewidth]{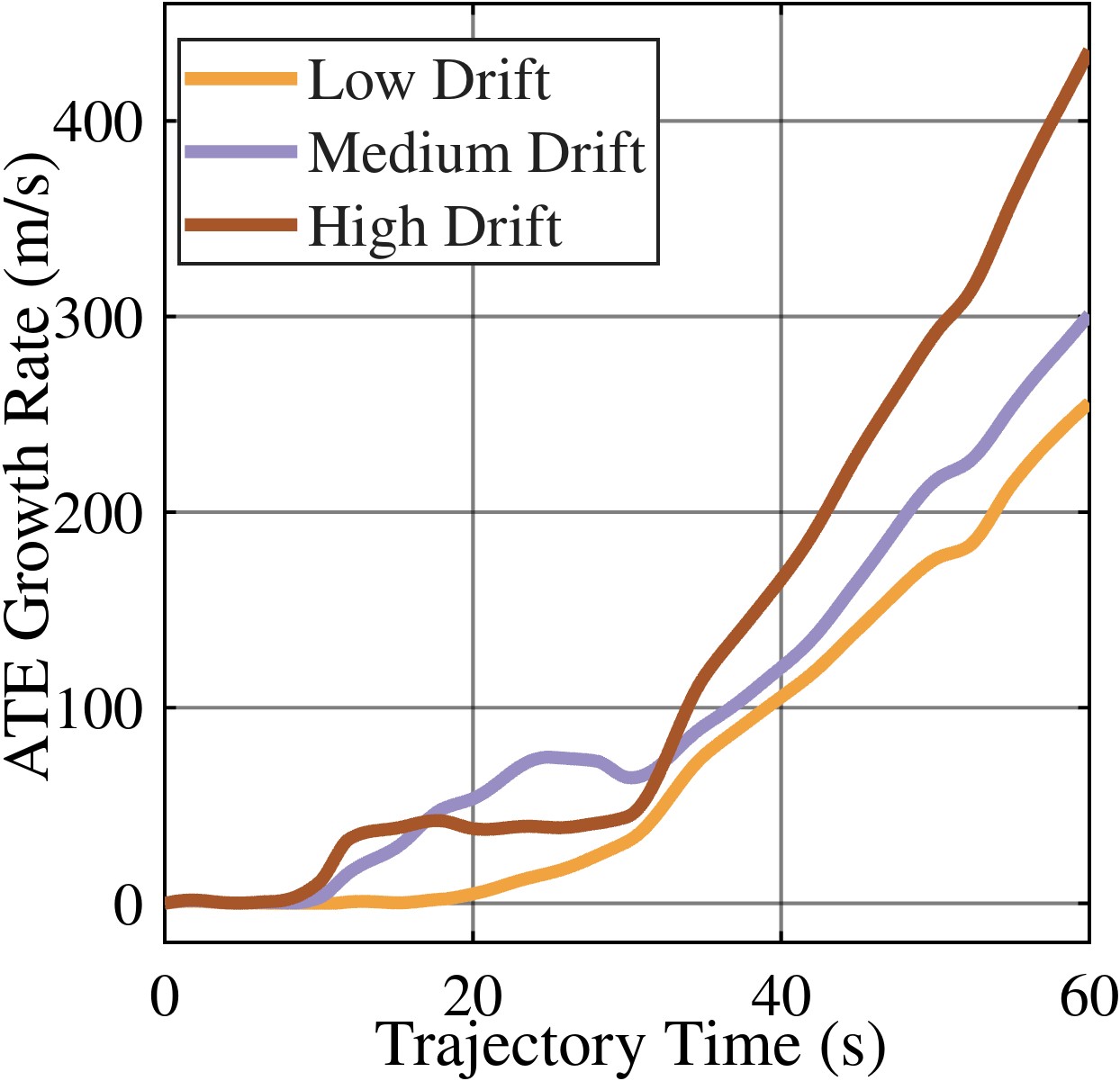}
        \caption{Translational error growth rate under IMU bias drift.}
        \label{fig:imu_bias_error_growth}
    \end{subfigure}
    \caption{Impact of IMU bias drift on VIO tracking. (a) Cumulative translational ATE and RPE for low, medium, and high drift levels. (b) Translational error growth rate showing accelerating divergence under sustained bias drift.}
    \label{fig:imu_bias_drift}
\end{figure}




We evaluate VIO robustness to gradual IMU calibration degradation by simulating gyroscope bias drift, which introduces a slowly accumulating systematic error in inertial measurements over time. Three drift severity levels are evaluated—low, medium, and high—representing realistic sensor aging or temperature-induced calibration drift during prolonged XR operation. Bias drift is applied throughout the trajectory to emulate persistent calibration instability. \cref{fig:imu_bias_ate_rpe} presents a stacked bar chart showing mean translational ATE and mean translational RPE for each drift level, with both metrics exhibiting catastrophic tracking failure. Even low drift produces a mean translational ATE of $932.33\,\mathrm{m}$ and a mean translational RPE of $37.80\,\mathrm{m}$. Medium drift increases the mean translational ATE to $1598.98\,\mathrm{m}$ ($71\%$ increase) and the mean translational RPE to $56.76\,\mathrm{m}$, while high drift further escalates errors to a mean translational ATE of $1826.83\,\mathrm{m}$ and a mean translational RPE of $67.53\,\mathrm{m}$.

\cref{fig:imu_bias_error_growth} shows the temporal evolution of the position error growth rate for all three drift levels, measured as the time derivative of translational ATE. This analysis reveals increasing growth rates over the trajectory, reflecting the compounding effect of orientation error introduced by gyroscope bias drift. Mean growth rates over the full trajectory reach $73.54\,\mathrm{m/s}$ for low drift, $100.38\,\mathrm{m/s}$ for medium drift, and $128.18\,\mathrm{m/s}$ for high drift, while peak instantaneous growth rates reach $253.48\,\mathrm{m/s}$, $298.83\,\mathrm{m/s}$, and $435.80\,\mathrm{m/s}$, respectively. These results show that bias drift leads to accelerating and unbounded position error accumulation as drift magnitude increases.

\subsection{Effect of Combined Camera and IMU Faults}

We evaluate VIO behavior under simultaneous camera and IMU faults by combining a 10\% random camera frame drop with 5$\times$ IMU noise amplification. This experiment compares three conditions: camera fault only, IMU fault only, and the combined fault case.

\begin{figure}[!t]
    \centering
    \includegraphics[width=\columnwidth]{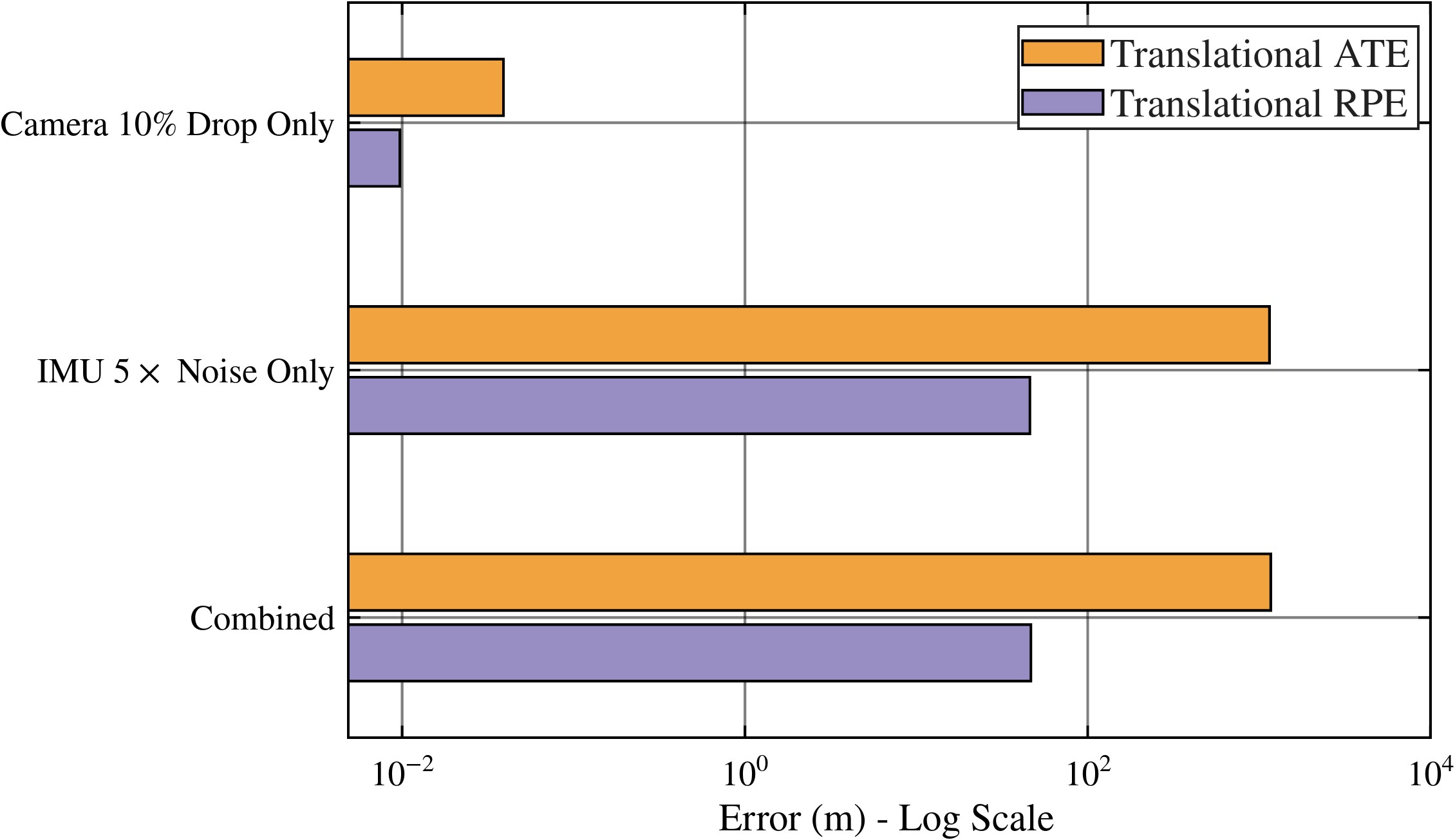}
    \caption{Mean Translational ATE and RPE under camera-only, IMU-only, and combined fault conditions. A logarithmic scale is used to visualize error magnitudes spanning centimeters to kilometers.}
    \label{fig:combined_faults}
\end{figure}

\cref{fig:combined_faults} reports mean translational ATE and RPE for all three conditions using a logarithmic scale. Camera frame drops alone produce minimal error, with an ATE of 3.90 cm and an RPE of 0.97 cm. IMU noise amplification alone results in catastrophic tracking failure, yielding ATE of 1150.38 m and RPE of 46.01 m. When both faults are applied simultaneously, the resulting error increases slightly to 1169.94 m ATE and 46.62 m RPE.

The combined fault case exceeds the sum of the individual errors by 19.52 m (1.7\%), indicating a weak positive interaction between the two fault types. Across all metrics, the error magnitude in the combined case remains within 2\% of that in the IMU-only fault condition. 

\section{Conclusion \& Discussion}
\label{sec:conclusion}

This work presented a controlled fault-injection study of visual-inertial odometry under camera and IMU degradation in XR-relevant settings. By systematically varying fault type, severity, timing, and duration, we directly compared failure modes across sensing modalities. \emph{The results reveal a strong asymmetry in VIO robustness}: IMU faults dominate system failure, producing errors several orders of magnitude larger than camera faults, while even severe visual degradation typically leads to bounded error growth. \emph{Under combined camera and IMU faults, error behavior is largely dictated by inertial degradation}, with visual faults contributing negligibly once inertial propagation is corrupted.

We further show that the \emph{temporal structure of faults plays a critical role in determining final accuracy}. Earlier faults lead to larger errors due to longer propagation horizons, and concentrated outages are more damaging than distributed losses, even when total fault duration is similar. \emph{IMU dropouts exhibit non-monotonic effects with respect to outage length}, indicating that estimator recovery dynamics, rather than duration alone, govern fault severity. Across both sensing modalities, \emph{corrupted measurements are more harmful than missing data}: noisy visual inputs mislead correction steps, while IMU noise and, in particular, bias drift directly corrupt the motion model. \emph{Bias drift introduces systematic orientation errors that compound through integration, in contrast to random noise}, which primarily increases stochastic dispersion and can be partially mitigated through sensor fusion.

These findings suggest that \emph{XR systems may benefit from treating inertial degradation in a gradual, confidence-aware manner rather than as a binary failure}, and in some cases prioritizing inertial robustness over visual fidelity. This study is limited to a single dataset, a single VIO implementation, and idealized fault models. Future work will extend this framework to additional datasets, alternative VIO architectures, longer trajectories, and more realistic fault traces, and will also explore adaptive fault mitigation and perceptual impact in immersive XR scenarios.

\section*{Supplemental Materials}
\label{sec:supplemental_materials}

All supplemental materials are available on GitHub at \url{https://github.com/Sourya17/industrial-vio-fault-benchmark}, released under a CC BY 4.0 license.

\bibliographystyle{abbrv-doi-hyperref-narrow}

\bibliography{template}
\end{document}